\documentclass[letterpaper, 10 pt, conference]{ieeeconf}  

\usepackage[utf8]{inputenc}
\usepackage{graphicx}
\usepackage{epstopdf}
\usepackage{etoolbox}
\usepackage{cite}
\usepackage{picinpar}
\usepackage{amsmath}
\usepackage{url}
\usepackage{flushend}
\usepackage{textcomp}
\usepackage{bm}
\usepackage{xcolor}
\usepackage{textcomp}
\usepackage{comment}
\usepackage{makecell}

\IEEEoverridecommandlockouts                              

\title{\LARGE \bf
Nonlinear Modeling for Soft Pneumatic Actuators\\ via Data-Driven Parameter Estimation 
}

\author{Wu-Te Yang$^{1}$, Hannah S. Stuart$^{2}$, Burak Kürkçü$^{3}$, and Masayoshi Tomizuka$^{1}$
\thanks{$^{1}$The authors are with the Department of Mechanical Engineering,
        University of California, Berkeley, MSC Lab, USA
        {\tt\small wtyang@berkeley.edu; tomizuka@berkeley.edu}}%
\thanks{$^{2}$The author is with the Department of Mechanical Engineering,
        University of California, Berkeley, Embodied Dexterity Lab, USA
        {\tt\small hstuart@berkeley.edu}}%
\thanks{$^{3}$The author is with the Department of Computer Engineering, Hacettepe University Turkey, and also is a Research Scholar at the University of California, Berkeley, USA
        {\tt\small bkurkcu@berkeley.edu}}%
}

\begin{document}

\maketitle
\thispagestyle{empty}
\pagestyle{empty}

\begin{abstract}

Precise modeling soft robots remains a challenge due to their infinite-dimensional nature governed by partial differential equations. This paper introduces an innovative approach for modeling soft pneumatic actuators, employing a nonlinear framework through data-driven parameter estimation. The research begins by introducing Ludwick's Law, providing a accurate representation of the large deflections exhibited by soft materials. Three key material properties, namely Young's modulus, tensile stress, and mixed viscosity, are utilized to estimate the parameters inside the nonlinear model using the least squares method. Subsequently, a nonlinear dynamic model for soft actuators is constructed by applying Ludwick's Law. To validate the accuracy and effectiveness of the proposed method, several experiments are performed demonstrating the model's capabilities in predicting the dynamic behavior of soft pneumatic actuators. In conclusion, this work contributes to the advancement of soft pneumatic actuator modeling that represents their nonlinear behavior.

\end{abstract}

\section{INTRODUCTION}

Soft robots have gained popularity and demonstrated superior performance and adaptability compared to traditional robots, especially but not limited to scenarios such as working in unknown environments~\cite{c1, Tang2023unknown}, delivering delicate components in the medical industry~\cite{alici2018bending}, and handling fragile objects in the food industry~\cite{c4, dai2023soft}. These soft robots are typically powered by various actuators, and soft pneumatic ones are the most widely used option~\cite{c5, c6} due to their lightweight, cost-effectiveness, and high power. However, partial differential equations (PDEs) governed dynamics of soft robots are highly nonlinear\cite{armanini2023soft}. Therefore, together with the complex nature of soft materials, identifying mechanical properties and representing the dynamics of soft robots are still challenging.

The stress-strain curve of soft materials can be described by Hooke's Law~\cite{c7, c20}. The theory is valid when soft materials are under limited strains~\cite{c30}, but the predicted errors increase in higher deformations. Several hyperelastic theories are presented to address the nonlinearity of soft materials~\cite{c25, c8}. Hyperelastic theories show higher accuracy than Hooke's Law, especially in large deformations. Although parameters of hyperelastic models are provided in a recent work~\cite{c25}, hyperelastic theories complicate the kinematic or dynamic modeling of soft robots. In~\cite{c22, c23, Brojan2009large}, Ludwick's Law equips the elongation term in Hooke's Law with a fractional power, making it applicable to high deformations. Ludwick's Law improves accuracy; however, to the best of the authors' knowledge, there is no systematic approach for determining the fractional power of the model.

\begin{figure}[t]
    \centering
    \includegraphics[width=220pt]{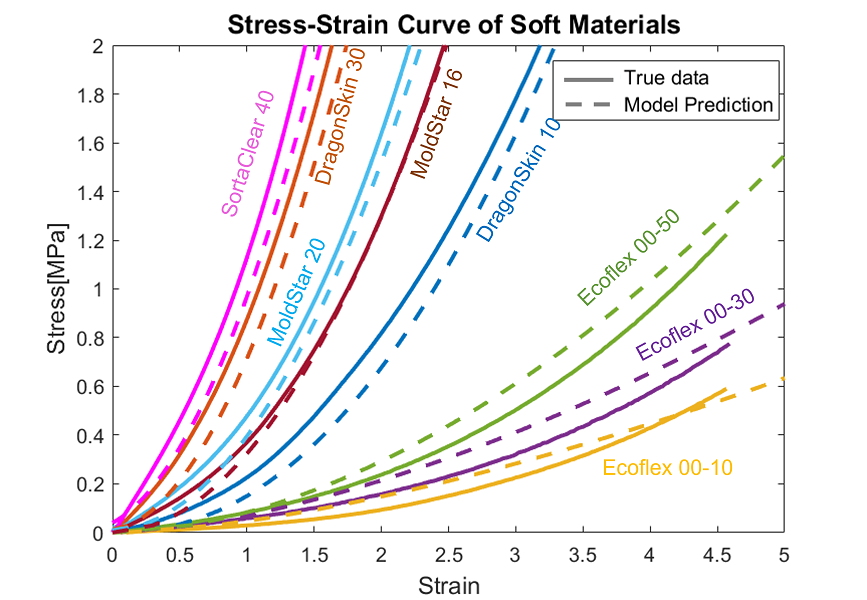}
    \caption{Stress-strain curve of soft materials (solid lines)~\cite{c25} versus predicted stress-strain curve of Ludwick's Law~\cite{c22} (dashed lines).}
    \label{fig: 1}
\end{figure}

Dynamical modeling of soft robots is a common topic, including the construction of bending models for soft actuators using methods such as the piece-wise constant curvature approach~\cite{c14}, Cosserat rod model~\cite{c15}, and Lagrange equation~\cite{c16}. While these methods yield accuracy, they entail complex modeling processes. Recent studies~\cite{c9, c10, c11, c12} have proposed modeling soft pneumatic systems as second-order dynamic systems, determining damping ratios and natural frequencies through response fitting. Second-order equations offer simplicity and accuracy at lower bending angles but become less accurate at higher angles. Combining Euler's bending beam theory~\cite{c18} and Ludwick's Law~\cite{c22} may provide a feasible, useful, and accurate dynamic model.

This paper proposes a nonlinear modeling of soft pneumatic actuators and presents a data-driven method to estimate the parameters within the model. Firstly, Ludwick's Law whose elongation term has fractional power is introduced. The fractional power is found to be influenced by material properties such as tensile stress, Young's modulus, and mixed viscosity. Those properties are used to build a least squares model for fractional power estimations. The theory is further implemented to build a nonlinear dynamic model for two soft pneumatic actuators. The nonlinear model is proposed to improve accuracy for some soft materials. The responses generated by the nonlinear dynamic equation for two actuators closely match the actual experimental responses. In summary, the proposed method serves as an alternative modeling approach for soft pneumatic actuators.

To understand the contributions of this paper in a comparative manner, related works are discussed. Lee and Brojan~\cite{c22,Brojan2009large} proposed the idea of using Ludwick's Law to model large deflection components. In this work, we extend that idea by proposing an approach to estimate the fractional power for various soft materials. Beda~\cite{c31} utilized a mathematical method to estimate parameters in hyperelastic models. By contrast, this research aims to use material properties to estimate a parameter within a nonlinear model. Porte et al.~\cite{c30} studied the influences of temperature and humidity on material properties of soft materials. In our work, we explore the influences of material properties on the nonlinearity and deformability of soft materials. Xaview~\cite{c9} and Wang~\cite{c11} modeled a soft robotic system as a second-order system, and the system parameters are obtained by using curve fitting. However, our work implemented a second-order nonlinear dynamic equation and the parameters are estimated by material properties.
To conclude, we seek to study the stress-strain relationship of soft materials and provide an alternative and functional modeling approach for soft robots.

The remainder of this paper is organized as follows. Section 2 formulates the theories of mechanical properties of materials. Section 3 discusses the dynamical modeling of soft robots. Section 4 demonstrates the experimental results, and Section 5 discusses and concludes the work.

\section{Theoretical Formulation}
This section introduces Hooke's Law (linear model) and Ludwick's Law (nonlinear model). The relationship between the nonlinear model and material properties is discussed. Those properties are also utilized to build a least-square model which is able to estimate the fractional power within Ludwick's Law.

\subsection{Stress-Strain Relationship}
Generally, the stress-strain curve of materials is described by Hooke's Law~\cite{c20}.
\begin{align}
    \begin{split}
    \sigma = E \epsilon
    \label{eqn: 1}
    \end{split}
\end{align}
where $\sigma = P/A$ is the stress and is defined as the applied force $P$ divided by the cross-sectional area $A$, $E$ is the Young's modulus, and $\epsilon = \frac{L-L_0}{L_0}$ is the strain and is defined as elongated length $L$ minus initial length $L_0$ divided by initial length.

\begin{figure}[t]
    \centering
    \includegraphics[width=220pt]{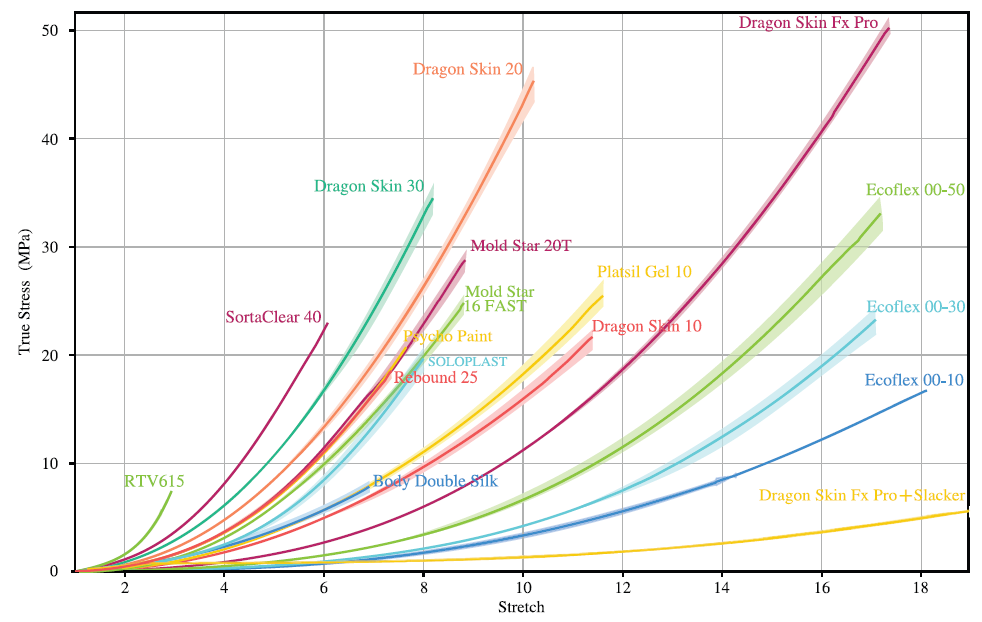}
    \caption{The stress-strain curve of commonly used soft materials are nonlinear~\cite{c25}.}
    \label{fig: 2}
\end{figure}

Equation~(\ref{eqn: 1}) depicts the linear relationship between stress and strain. This equation is valid under small elongations or when it is applied to hard materials such as metals~\cite{c20}. However, soft materials usually exhibit large elongations because of relatively small Young's modulus~\cite{c25}. The nonlinear stress-strain curve of common soft materials are shown in Fig.~\ref{fig: 2}. The linear stress-strain curve is no longer valid since the curve is nonlinear~\cite{c21}. As the materials are under large deformations, Ludwick's relation should be applied~\cite{c22,c23,Brojan2009large} 
\begin{align}
    \begin{split}
    \sigma = E {\epsilon}^{n}
    \label{eqn: 2}
    \end{split}
\end{align}
where $n \in R$ is a fractional number and it varies with materials.

Based on Eq.~(\ref{eqn: 2}), the stress-strain curve is nonlinear and the power $n$ is dependent on the properties of materials. The next issue will be proposing a systematic way to determine the fractional power $n$. In contrast to the limited options available for integer values, finding an appropriate $n$ as a fractional number poses a time-consuming task, often requiring trial-and-error-based approaches.

\subsection{Fractional Power and Material Properties}
\label{FNMP}
The $n$ in Eq.~(\ref{eqn: 2}) is usually obtained empirically or experimentally. Thus, this subsection aims to explore the relationship between material properties and the fractional power $n$ of Eq.~(\ref{eqn: 2}). With the relationship between $n$ and material properties, the fractional power can be determined and Eq.~(\ref{eqn: 2}) has the potential to be applied in future research.

When it comes to material properties, there are a large number of choices such as density, heat capacity, coefficient of thermal expansion, etc. However, we could pick key material properties according to Eq.~(\ref{eqn: 2}). Tensile strength, Young's modulus, and mixed viscosity are selected to study their relationship with the fractional power $n$. Tensile strength represents the maximum external force, $\sigma$, that a material is capable of resisting. Young's modulus characterizes its stiffness and ability to withstand stretching. Mixed viscosity quantifies the material's resistance to flow when soft materials are in a liquid state. The commonly used soft materials turn into a solid state by mixing liquid rubber with curing agents~\cite{c21}. The effect of viscosity, thus, might influence the elongation term ($\epsilon$)~\cite{c24}. In Fig.~\ref{fig: 3}, each property shows a correlation with the fractional number $n$. 

This research studies 10 commonly used soft materials, namely Dragon Skin FX-Pro, Dragon Skin\texttrademark 10 MEDIUM, Dragon Skin 20, Dragon Skin 30, Ecoflex\texttrademark 00-10, Ecoflex 00-30, Ecoflex 00-50, Mold Star\texttrademark 16 FAST, Mold Star 20T, and SORTA-Clear\texttrademark 40. Their properties are also provided on the supplier's website~\cite{c26}. Their stress-strain curves are provided in the online library~\cite{c25}. The fractional power of each soft material is obtained by curve fitting the stress-strain curve in the library. The properties and fractional numbers of the materials are displayed in Table I.

\begin{table}[ht]
\caption{Fractional power and material properties of selected soft materials~\cite{c26}}
\label{tab:my_label}
\centering
\scalebox{0.95}{
\begin{tabular}{|c|c|c|c|c|}
\hline
\makecell{\textbf{Material}} & \textbf{Fractional} & \textbf{Young's} & \textbf{Mixed} & \textbf{Tensile} \\
\textbf{name} & \textbf{power} & \textbf{modulus} & \textbf{viscosity} & \textbf{strength}\\
& \textbf{(n)} & \textbf{(MPa)} & \textbf{(cps)} & \textbf{(MPa)}\\
\hline
\makecell{Dragon Skin} & & & &\\
FX Pro & 1.538 & 0.26 & 18000 & 1.99\\ 
\hline
\makecell{Dragon Skin\texttrademark} & & & &\\
10 MEDIUM & 2.174 & 0.15 & 23000 & 3.28\\ 
\hline
\makecell{Dragon Skin} & & & &\\
20 & 2.500 & 0.34 & 20000 & 3.79\\ 
\hline
\makecell{Dragon Skin} & & & &\\
30 & 2.222 & 0.59 & 20000 & 3.45\\ 
\hline
\makecell{Ecoflex\texttrademark} & & & &\\
00-10 & 1.538 & 0.06 & 14000 & 0.83\\ 
\hline
\makecell{Ecoflex} & & & &\\
00-30 & 1.613 & 0.07 & 3000 & 1.38\\ 
\hline
\makecell{Ecoflex} & & & &\\
00-50 & 1.818 & 0.08 & 8000 & 2.17\\ 
\hline
\makecell{Mold Star\texttrademark} & & & &\\
16 FAST & 2.000 & 0.38 & 12500 & 2.76\\ 
\hline
\makecell{Mold Star} & & & &\\
20T & 2.174 & 0.32 & 11000 & 2.90\\ 
\hline
\makecell{SORTA-Clear\texttrademark} &  &  &  &\\
40 & 2.500 & 0.62 & 35000 & 5.51\\ 
\hline
\end{tabular}} 
\end{table}

\subsection{Fractional Power Estimation}
\label{FNE}
\begin{figure}[t]
    \centering
    \includegraphics[width=220pt]{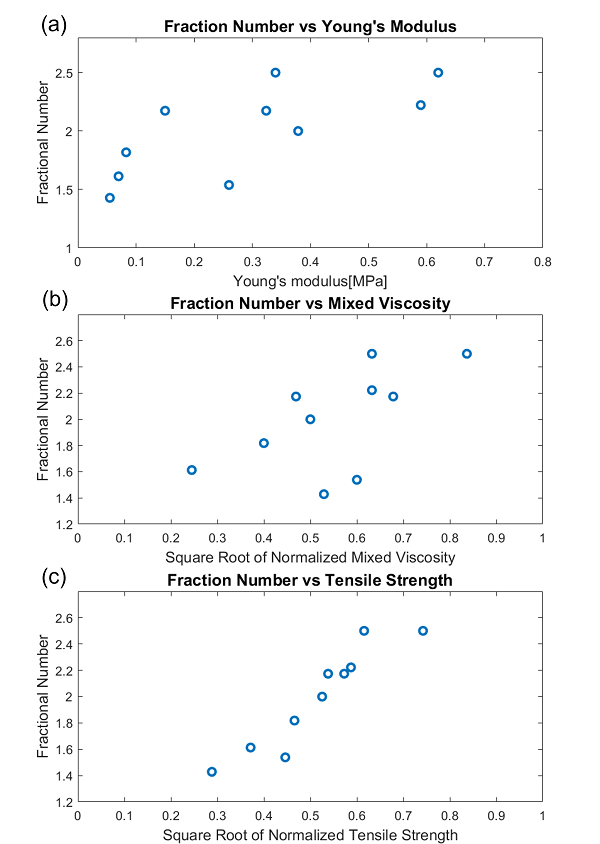}
    \caption{Selected material properties correlate to the fractional power $n$.}
    \label{fig: 3}
\end{figure}

These three properties encapsulate the intricate behavior of a material under stress as displayed in Fig.~\ref{fig: 3}. To harness these findings effectively, we employ the technique of least squares regression to construct a prediction model. Those selected properties serve as predictors and the fractional power acts as a response. By doing so, the model mathematically captures the relationships between these properties and the fractional power, enabling them to make accurate forecasts about a material's performance under various stresses, which is invaluable in the field of soft robotics. Thus, we can utilize a model to predict the stress-strain curve (fractional number) when properties of a soft material are given.

The standard least-square model~\cite{c27} is described as
\begin{align}
    \begin{split}
    y = Ax
    \label{eqn: 3}
    \end{split}
\end{align}
where $A \in R^{m\times n}$ is a matrix that contains the predictors of each experiment, $y \in R^m$ is a column vector that includes the measured responses, and $x \in R^n$ is a column vector whose parameters are to be solved. The $m$ represents the data number, while $n$ is the number of predictors.

The closed form solution of Eq.~(\ref{eqn: 3}) is obtained by pre-multiplying $A^T$ on both side and then pre-multiplying inverse of ${A^T}A$
\begin{align}
    \begin{split}
    {A^T}Ax = {A^T}y\\
    x = {({A^T}A)^{-1}}{A^T}y
    \label{eqn: 4}
    \end{split}
\end{align}
where $({A^T}A)^{-1}$ is nonsingular, invertible, and exists.

Based on Eq.~(\ref{eqn: 4}). the $y$ vector contains the fractional power of each selected material in Table I of Sec.~\ref{FNMP}
\begin{align}
    \begin{split}
    y = \begin{bmatrix}
           n_{1} & n_{2} & \hdots & n_{m}
         \end{bmatrix}^{T}
    \label{eqn: 5}
    \end{split}
\end{align}
and the $x$ vector contains the parameters to be solved and it is depicted as
\begin{align}
    \begin{split}
    x = \begin{bmatrix}
           x_{1} & x_{2} & x_{3}
         \end{bmatrix}^T
    \label{eqn: 6}
    \end{split}
\end{align}
and the $A$ matrix includes the material properties in Table I of Sec.~\ref{FNMP}
\begin{align}
    \begin{split}
    A = \begin{bmatrix}
        E_1 & {MV_1}^{0.5} & {TS_1}^{0.5}\\
        E_2 & {MV_2}^{0.5} & {TS_2}^{0.5}\\
        & \vdots & \\
        E_m & {MV_m}^{0.5} & {TS_m}^{0.5}
        \end{bmatrix}
    \label{eqn: 7}
    \end{split}
\end{align}
where $MV$ is the abbreviation of the mixed viscosity which is normalized by 50000 $cps$, and $TS$ is the abbreviation of the tensile strength which is normalized by 10 $MPa$. Since 10 soft materials are selected in Sec.~\ref{FNMP}, they are divided into 8 and 2. The 8 soft materials are used to build the least squares model. Thus, the $m$ is 8 here and the other 2 materials (Dragon Skin FX-Pro and Dragon Skin 20) will be used to verify the least squares model.

With Eq.~(\ref{eqn: 5})-(\ref{eqn: 7}), we get the values of $x$ vector by using Eq.~(\ref{eqn: 4}), and least-square model is represented as
\begin{align}
    \begin{split}
    y = E{x_1} + MS^{0.5}{x_2} + TS^{0.5}{x_3}
    \label{eqn: 8}
    \end{split}
\end{align}
This model will be used to predict the fractional number ($n$) if given new or unknown soft materials. The results will be displayed in Sec.~\ref{exp}.

\begin{figure}[t]
    \centering
    \includegraphics[width=190pt]{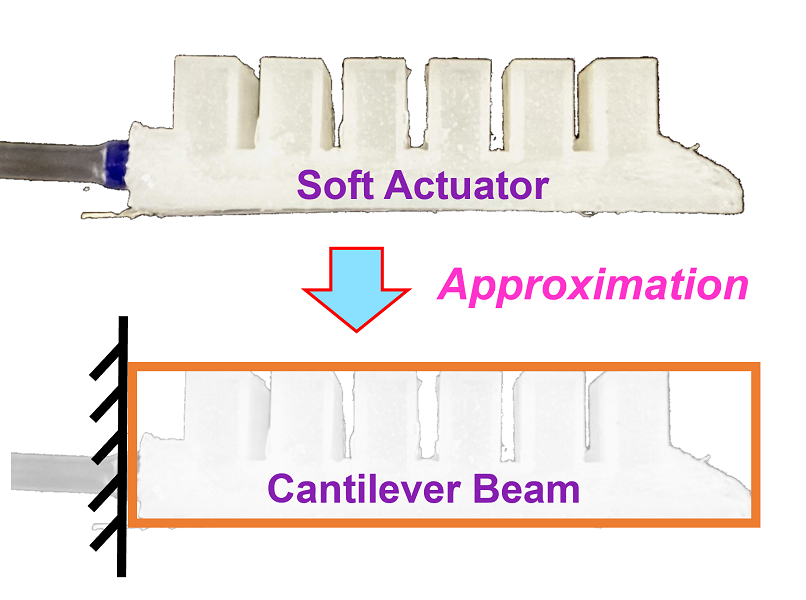}
    \caption{The irregular geometric structure of the soft pneumatic actuator is approximated by a cantilever beam.}
    \label{fig: 4}
\end{figure}

\section{System Modeling}
Obtaining the parameter within Ludwick Law, Eq.~(\ref{eqn: 2}) is used to build the dynamic model of soft robotic systems. Here, we will focus on modeling soft pneumatic actuators(SPAs) in our previous work~\cite{c19}. The MATLAB\textregistered simulation results are also demonstrated in this section.

SPAs have irregular geometric shapes since they consist of several discrete chambers. The nonlinear structure is approximated as a cantilever beam. The simplified structure makes the dynamic analysis possible. The analytical methods for a cantilever beam can be applied~\cite{c19}. Instead of using Hooke's Law (\ref{eqn: 1}), we apply Ludwick's Law (\ref{eqn: 2}) and the analytical method for a cantilever beam.

The approximated beam structure is shown in Fig.~\ref{fig: 4}. If a force is applied at the free end of the beam, the structure will bend and generate a bending angle. The bending angle can be depicted as~\cite{c20}
\begin{align}
    \begin{split}
    P = (\frac{2EI}{L_0^2}){\theta} = K \theta 
    \label{eqn: 9}
    \end{split}
\end{align}
\begin{align}
    \begin{split}
    K = \frac{2EI}{L_0^2} 
    \label{eqn: 9_1}
    \end{split}
\end{align}
where $F$ is the force acted at the free end and the force here is generated by the pressure, $L_0$ is the initial length of the structure, $K$ is the spring constant, $\theta$ is the bending angle, and $I$ is the moment of inertia which depicted as
\begin{align}
    \begin{split}
    {I} = (\frac{1}{12})b{h}^{3} 
    \label{eqn: 10}
    \end{split}
\end{align}

\begin{figure}[t]
    \centering
    \includegraphics[width=160pt]{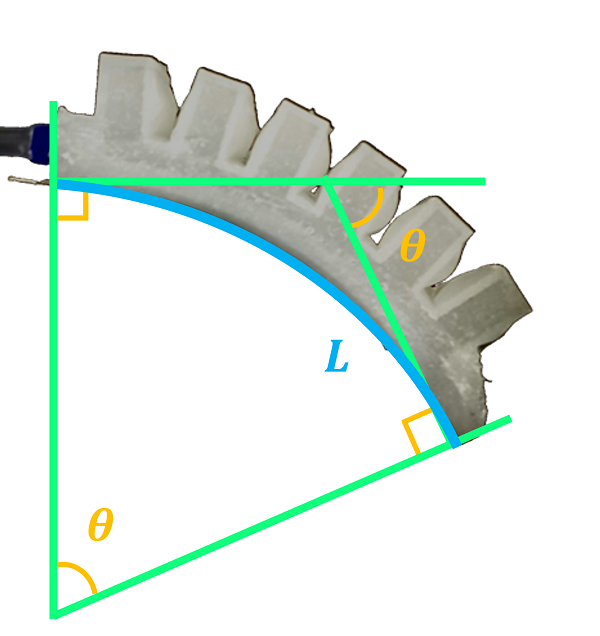}
    \caption{The bending geometric of the soft actuator when it is pressurized.}
    \label{fig: 5}
\end{figure}

Equation~(\ref{eqn: 9}), however, is derived based on linear model assumption~\cite{c20}. It may not describe the dynamics of nonlinear soft materials. To apply the nonlinear model as Eq.~(\ref{eqn: 2}), the model is adjusted according to~\cite{c22}. When the pressure is applied to the soft actuator, it will bend as shown in Fig.~\ref{fig: 5}. The bending angle can be described as

\begin{align}
    \begin{split}
    P = (\frac{n+1}{n})^n(\frac{EI_n}{L_0^{n+1}}){\theta}^n = K_n {\theta}^n
    \label{eqn: 11}
    \end{split}
\end{align}
\begin{align}
    \begin{split}
    K_n = (\frac{n+1}{n})^n(\frac{EI_n}{L_0^{n+1}})
    \label{eqn: 11_1}
    \end{split}
\end{align}
where $K_n$ is the spring constant when $n>1$, $I_n$ is the modified moment of inertia for a large deflection component, and it is displayed as

\begin{align}
    \begin{split}
    {I_n} = (\frac{1}{2})^n(\frac{1}{2+n})b{h}^{(2+n)} 
    \label{eqn: 12}
    \end{split}
\end{align}

If $n=1$, Eq.~(\ref{eqn: 11}), (\ref{eqn: 11_1}), and (\ref{eqn: 12}) degrade to Eq.~(\ref{eqn: 9}), (\ref{eqn: 9_1}), and (\ref{eqn: 10}) and the model becomes linear.

With Eq.~(\ref{eqn: 11}), the bending dynamic equation of the soft pneumatic actuator is built by 
\begin{align}
    \begin{split}
    F - C_n{\dot \theta} - K_n {\theta}^n = M \ddot{\theta}
    \label{eqn: 13}
    \end{split}
\end{align}
After manipulation, the nonlinear dynamic model of the soft actuator is shown as
\begin{align}
    \begin{split}
    M \ddot{\theta} + C_n{\dot \theta} + K_n {\theta}^n = F
    \label{eqn: 14}
    \end{split}
\end{align}
where $M$ is the mass of the soft actuator, $C_n$ is the damper of the soft actuator, $K_n$ is the spring constant obtained from (\ref{eqn: 11_1}). Mass is measured by weight scale and $C_n$ currently is estimated by curve fitting.

The dynamic equation is similar to the second-order equation, but it has a nonlinear spring term based on Ludwick's Law, Eq.~(\ref{eqn: 2}). The fractional power ($n$) can be estimated by using Eq.~(\ref{eqn: 8}) and the selected material properties.

\section{Experimental Evaluation}
\label{exp}

\subsection{Soft Actuators Setup}
In Sec.~\ref{FNE}, eight materials are used to build a least squares model and the other two will be used to verify. Thus, two soft actuator prototypes are made of Ecoflex\textregistered Dragon Skin FX-Pro and Ecoflex\textregistered Dragon Skin 20 respectively.
Two actuators have upper and bottom components which are fabricated by using two different molds~\cite{c19}. They are then bonded together by the silicone adhesive, Sil-poxy\textregistered. The nozzle at their end is connected to the syringe pump~\cite{yang2023pump} to provide air pressure. The bottom component has a piece of flex sensor embedded inside~\cite{c29}. Two soft actuators have the same dimensions.
 
The estimated fractional power of Dragon Skin 20 and Dragon Skin-FX Pro are 2.365 and 1.727 respectively as displayed in Table II. The estimated fractional powers are close to the true values but have limited errors. With the fractional powers, the dynamic equation of soft actuators is setup.

\begin{figure}[t]
    \centering
    \includegraphics[width=210pt]{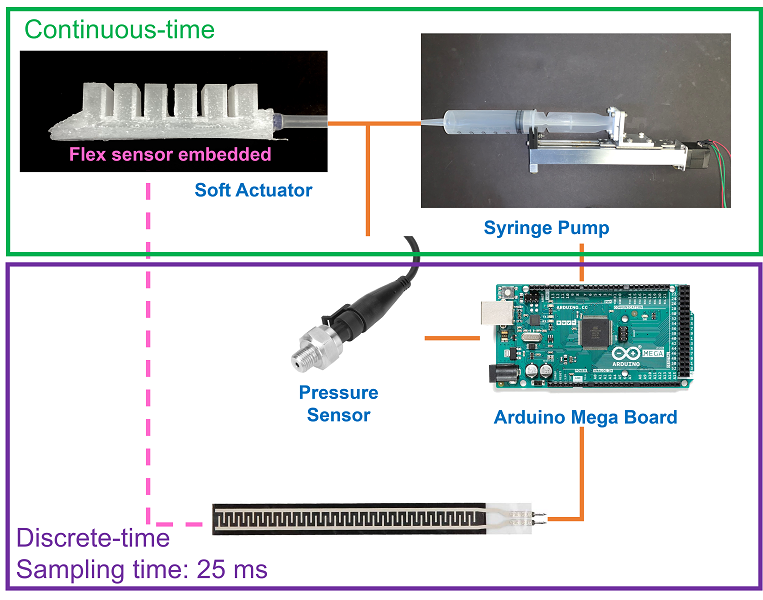}
    \caption{The schematic of the experimental setup.}
    \label{fig: 7}
    \vspace{-0.15in}
\end{figure}

\begin{table}[http] 
\centering
\caption{\label{tab:Table II}Comparison of estimated and true fractional powers}
\begin{tabular}{|c | c | c|}
\hline
\textbf{Material Name} & \textbf{True Value} & \textbf{Estimated Value}\\ [0.5ex]
\hline
Dragon Skin 20 & 2.174 & 2.365 \\ [0.5ex]
\hline
Dragon Skin FX-Pro & 1.538  & 1.727\\ [0.5ex] 
\hline
\end{tabular}
\end{table}

\subsection{Experimental Setup}

Figure~\ref{fig: 7} presents the control block diagram and the experimental setup. Soft actuators are powered by a custom-designed syringe pump~\cite{yang2023pump}. To facilitate open-loop control, an air pressure sensor (Walfront in Lewes, DE) is employed, offering a sensing range of 0 to 80 psi to monitor air pressure. Within the actuator, a flex sensor (Walfront in Lewes, DE) is integrated to measure the bending angle, enabling feedback control. Both sensors are synchronized with the Arduino MEGA 2560 microcontroller (SparkFun Electronics, Niwot, CO). This microcontroller is based on the Microchip ATmega 2560 platform. Furthermore, the microcontroller is connected to a computer to record the sensing data.

\begin{figure}[t]
    \centering
    \includegraphics[width=220pt]{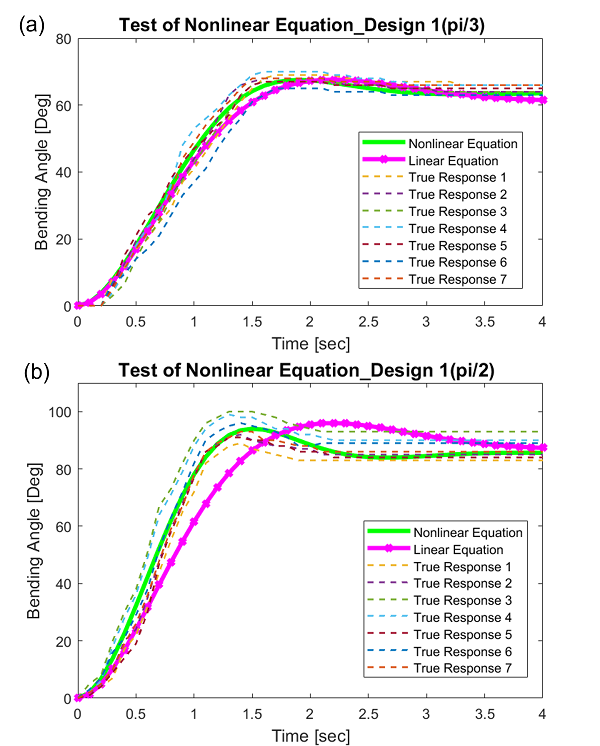}
    \caption{Comparisons between seven experimental data, linear dynamical equation and nonlinear dynamical equation of the soft actuator made of Dragon Skin 20.}
    \label{fig: 8}
    \vspace{-0.15in}
\end{figure}

\subsection{Step Response Test}

Step response tests intend to evaluate the proposed nonlinear model. Equation~(\ref{eqn: 14}) is solved by MATLAB solver ``ode45'' to get the system responses. The schematic of the bending angle of soft actuators is shown in Fig.~\ref{fig: 5}. The bending angles are measured by the flex sensor embedded inside the soft actuators. In addition, there are two soft actuators made of Dragon Skin 20 and Dragon Skin FX-Pro. The former one is named ``Design 1'', while the latter is called ``Design 2''.

\subsubsection{Design 1 Test}The step response tests are visualized in Fig.~\ref{fig: 8}. In Fig.~\ref{fig: 8} (a), the setpoint is 60 degrees. Seven true responses are done and added in the figure to evaluate model's accuracy. Additionally, a linear second-order dynamic equation was proposed to model soft actuators in our previous work~\cite{yang2023control}. The linear equation is plotted in the same figure. In Fig.~\ref{fig: 8} (a), the nonlinear dynamic equation describes the system's responses better than the linear equation. The root-mean-square error(RMS) of the nonlinear equation and the average of true responses is 1.29 degrees (2.2\%), while the RMS of the linear equation and the average of true responses is 2.03 degrees (3.4\%). 

Another step response test is conducted with a different setpoint (90 degrees). We would like to evaluate the model's accuracy with a higher bending angle. The results are demonstrated in Fig.~\ref{fig: 8} (b). The RMS error between the nonlinear model and the average of true responses is 3.17 degrees (3.5\%), while the RMS error of the linear equation with true response is 6.15 degrees (6.9\%). The nonlinear dynamic equation has smaller errors; thus, the nonlinear dynamic equation is more accurate than linear dynamic equation.

\subsubsection{Design 2 Test}
\begin{figure}[t]
    \centering
    \includegraphics[width=220pt]{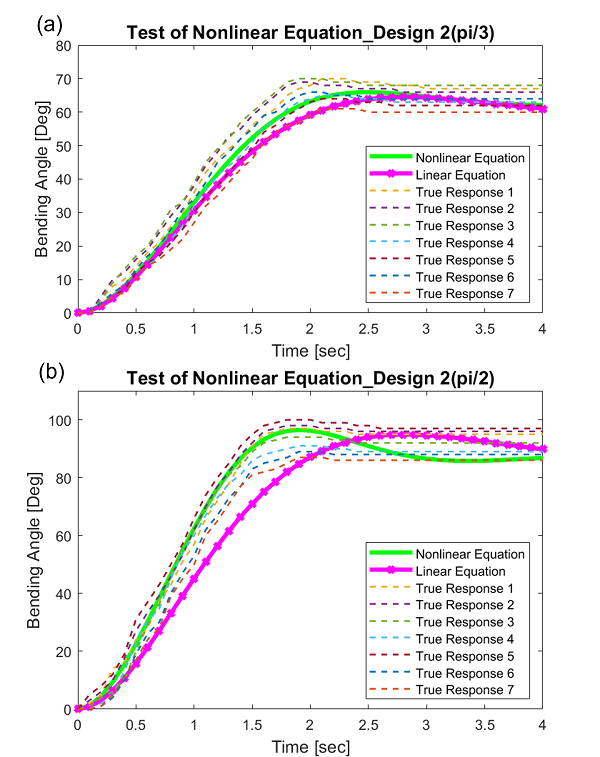}
    \caption{Comparisons between seven experimental data, linear dynamical equation and nonlinear dynamical equation of the soft actuator made of Dragon Skin FX-Pro.}
    \label{fig: 9}
    \vspace{-0.15in}
\end{figure}

The step response tests are visualized in Fig.~\ref{fig: 9}. The setpoint is 60 degrees in the first test (Fig.~\ref{fig: 9}(a)). The RMS error of the nonlinear equation and average of true responses is 1.46 degrees (2.4\%) and that of the linear equation and average of true responses is 3.49 degrees (5.8\%). The setpoint is 90 degrees in the second test (Fig.~\ref{fig: 9}(b)). The RMS errors between both equations and the average of true responses are 3.61 (4.0\%) and 7.98 degrees (8.9\%) respectively. Again, the nonlinear dynamic equation outperforms.

\section{Discussion and Conclusion}

\subsection{Discussions}
The results presented in Sec.~\ref{exp} demonstrate the validity of the linear second-order equation under conditions of limited bending angles. When the bending angle remains within a limited range (60 degrees), the elongation of the soft material remains relatively modest. This observation is corroborated by the stress-strain curve, where it is evident that Hooke's Law (\ref{eqn: 1}) closely approximates the experimental data. However, as the bending angle increases, so does the extent of elongation in the material. In such cases, the linear equation may no longer suffice, and it becomes necessary to consider the application of Ludwick's Law (\ref{eqn: 2}). 

Another benefit of using Eq.~(\ref{eqn: 2}) lies in the fractional power ($n$) of the nonlinear equation. Since soft materials have uncertainty and their behaviors vary with environmental conditions~\cite{c30}, Eq.~(\ref{eqn: 14}) could be modified as $M \ddot{\theta} + C_n {\dot \theta} + K_n {\theta}^{n+\Delta n} = F$. The $\Delta n$ represents the uncertainty of soft materials. The small perturbation term is added to finely adjust the dynamic model of soft robots. By contrast, if $\Delta n$ is an integer number ($\Delta n \in N$), it is hard to adjust the stress-strain curve. That is, the $n + \Delta n$ will be an integer such as 1, 2, 3, etc. The stress-strain curve changes obviously as the perturbation term is added, so the curve cannot catch true experimental data precisely. To sum up, the fractional power can be any real value and the perturbation term ($\Delta n$) is smaller than unity. Therefore, a nonlinear dynamic equation has more degree of freedom than an integer-order equation. 



\subsection{Conclusion}

This study introduces a novel nonlinear modeling approach for soft pneumatic actuators and proposes an effective method for estimating the parameters within the nonlinear model. Our approach involves describing the stress-strain curve of soft materials using Ludwick's Law, which accurately captures the true stress-strain data. Instead of relying on conventional curve fitting methods, we employ the least squares method to estimate the parameters within this nonlinear model.
To facilitate parameter estimation, we leverage key material properties such as tensile stress, Young's modulus, and mixed viscosity as predictors in our least squares model. Our approach yields predicted parameters that closely align with the true values, with an error margin of less than 10\%.
Furthermore, we extend the application of Ludwick's Law to formulate the dynamic equations for two soft actuator prototypes, each constructed from different soft materials. Experimental step response tests validate the accuracy and reliability of our nonlinear model. Specifically, we observe that the RMS errors of the nonlinear dynamic equations remain within 3 degrees even at higher bending angles. In contrast, the linear dynamic equation exhibits larger RMS errors of up to 8 degrees.
In conclusion, our innovative modeling methodology and parameter estimation approach offer an alternative for modeling soft materials.

\bibliographystyle{ieeetr}
\bibliography{IEEEabrv}

\begin{thebibliography}{10}

\bibitem{c1}
F.~Iida and C.~Laschi, ``Soft robotics: Challenges and perspectives,'' {\em
  Procedia Computer Science}, vol.~7, no.~1, pp.~99--102, 2011.

\bibitem{Tang2023unknown}
Z.~Tang, P.~Wang, W.~Xin, Z.~Xie, L.~Kan, M.~Mohanakrishnan, and C.~Laschi,
  ``Meta-learning-based optimal control for soft robotic manipulators to
  interact with unknown environments,'' in {\em IEEE International Conference
  on Robotics and Automation (ICRA)}, pp.~982--988, IEEE, 2023.

\bibitem{alici2018bending}
G.~Alici, T.~Canty, R.~Mutlu, W.~Hu, and V.~Sencadas, ``Modeling and
  experimental evaluation of bending behavior of soft pneumatic actuators made
  of discrete actuation chambers,'' {\em Soft Robotics}, vol.~5, no.~1,
  pp.~24--35, 2018.

\bibitem{c4}
E.~Navas, R.~Fernández, D.~Sepúlveda, M.~Armada, and P.~Gonzalez-de Santos,
  ``Soft grippers for automatic crop harvesting: A review,'' {\em Sensors},
  vol.~21, no.~8, p.~2689, 2021.

\bibitem{dai2023soft}
R.~Zhu, D.~Fan, W.~Wu, C.~He, G.~Xu, J.~S. Dai, and H.~Wang, ``Soft robots for
  cluttered environments based on origamianisotropic stiffness structure (oass)
  inspired by desertiguana,'' {\em Advanced Intelligent Systems}, p.~2200301,
  2023.

\bibitem{c5}
S.~Zaidi, M.~Maselli, C.~Laschi, and M.~Cianchetti, ``Actuation technologies
  for soft robot grippers and manipulators: A review,'' {\em Current Robotics
  Reports}, pp.~1--15, 2021.

\bibitem{c6}
J.~Hughes, U.~Culha, F.~Giardina, F.~Guenther, A.~Rosendo, and F.~Iida, ``Soft
  manipulators and grippers: a review,'' {\em Frontiers in Robotics and AI},
  vol.~3, p.~69, 2016.

\bibitem{armanini2023soft}
C.~Armanini, F.~Boyer, A.~T. Mathew, C.~Duriez, and F.~Renda, ``Soft robots
  modeling: A structured overview,'' {\em IEEE Transactions on Robotics}, 2023.

\bibitem{c7}
D.~Rus and M.~T. Tolley, ``Design, fabrication and control of soft robots,''
  {\em Nature}, vol.~521, no.~7553, pp.~467--476, 2015.

\bibitem{c20}
R.~C. Hibbeler, {\em Mechanics of materials 8th}.
\newblock Pearson, New York, 2017.

\bibitem{c30}
E.~Porte, S.~Eristoff, A.~Agrawala, and R.~Kramer-Bottiglio, ``Characterization
  of temperature and humidity dependence in soft elastomer behavior,'' {\em
  Soft Robotics}, 2023.

\bibitem{c25}
L.~Marechal, P.~Balland, L.~Lindenroth, F.~Petrou, C.~Kontovounisios, and
  F.~Bello, ``Toward a common framework and database of materials for soft
  robotics,'' {\em Soft robotics}, vol.~8, no.~3, pp.~284--297, 2021.

\bibitem{c8}
L.~Chen, C.~Yang, H.~Wang, J.~S. Bransonc, David T.and~Dai, and R.~Kang,
  ``Design and modeling of a soft robotic surface with hyperelastic
  material.,'' {\em Mechanism and Machine Theory}, vol.~130, pp.~109--122,
  2018.

\bibitem{c22}
K.~Lee, ``Large deflections of cantilever beams of non-linear elastic material
  under a combined loading,'' {\em International Journal of Non-Linear
  Mechanics}, vol.~37, no.~3, pp.~439--443, 2002.

\bibitem{c23}
S.~Ghuku and K.~N. Saha, ``A review on stress and deformation analysis of
  curved beams under large deflection,'' {\em International Journal of
  Engineering and Technologies}, vol.~11, pp.~13--39, 2017.

\bibitem{Brojan2009large}
M.~Brojan, T.~Videnic, and F.~Kosel, ``Large deflections of nonlinearly elastic
  non-prismatic cantilever beams made from materials obeying the generalized
  ludwick constitutive law.,'' {\em Meccanica}, vol.~44, pp.~733--739, 2009.

\bibitem{c14}
C.~Della~Santina, R.~K. Katzschmann, A.~Biechi, and D.~Rus, ``Dynamic control
  of soft robots interacting with the environment,'' in {\em IEEE International
  Conference on Soft Robotics (RoboSoft)}, pp.~46--53, IEEE, 2018.

\bibitem{c15}
A.~Doroudchi and S.~Berman, ``Configuration tracking for soft continuum robotic
  arms using inverse dynamic control of a cosserat rod model,'' in {\em IEEE
  International Conference on Soft Robotics (RoboSoft)}, pp.~207--214, IEEE,
  2021.

\bibitem{c16}
C.~M. Best, M.~T. Gillespie, P.~Hyatt, L.~Rupert, V.~Sherrod, and M.~D.
  Killpack, ``A new soft robot control method: Using model predictive control
  for a pneumatically actuated humanoid,'' {\em IEEE Robotics and Automation
  Magazine}, vol.~23, no.~3, pp.~75--84, 2016.

\bibitem{c9}
M.~S. Xavier, A.~J. Fleming, and Y.~K. Yong, ``Nonlinear estimation and control
  of bending soft pneumatic actuators using feedback linearization and ukf,''
  {\em IEEE/ASME Transactions on Mechatronics}, 2022.

\bibitem{c10}
E.~H. Skorina, M.~Luo, W.~Tao, F.~Chen, J.~Fu, and C.~D. Onal, ``Adapting to
  flexibility: model reference adaptive control of soft bending actuators,''
  {\em IEEE Robotics and Automation Letters}, vol.~2, no.~2, pp.~964--970,
  2017.

\bibitem{c11}
T.~Wang, Y.~Zhang, Z.~Chen, and S.~Zhu, ``Parameter identification and
  model-based nonlinear robust control of fluidic soft bending actuators,''
  {\em IEEE/ASME transactions on mechatronics}, vol.~24, no.~3, pp.~1346--1355,
  2019.

\bibitem{c12}
E.~H. Skorina, M.~Luo, S.~Ozel, F.~Chen, W.~Tao, and C.~D. Onal, ``Feedforward
  augmented sliding mode motion control of antagonistic soft pneumatic
  actuators,'' in {\em IEEE International Conference on Robotics and Automation
  (ICRA)}, pp.~2544--2549, IEEE, 2015.

\bibitem{c18}
K.~M. de~Payrebrune and O.~M. O’Reilly, ``On constitutive relations for a
  rod-based model of a pneu-net bending actuator,'' {\em Extreme Mechanics
  Letters}, vol.~9, pp.~38--46, 2016.

\bibitem{c31}
T.~Beda, ``An approach for hyperelastic model-building and parameters
  estimation a review of constitutive models.,'' {\em European Polymer
  Journal}, vol.~50, pp.~97--108, 2014.

\bibitem{c21}
M.~S. Xavier, A.~J. Fleming, and Y.~K. Yong, ``Finite element modeling of soft
  fluidic actuators: Overview and recent developments,'' {\em Advanced
  Intelligent Systems}, vol.~3, no.~2, p.~2000187, 2021.

\bibitem{c24}
R.~Valette, E.~Hachem, M.~Khalloufi, A.~Pereira, M.~Mackley, and S.~Butler,
  ``The effect of viscosity, yield stress, and surface tension on the
  deformation and breakup profiles of fluid filaments stretched at very high
  velocities,'' {\em Journal of Non-Newtonian Fluid Mechanics}, vol.~263,
  pp.~130--139, 2019.

\bibitem{c26}
``{Smooth-on/platinum-silicone}.''
  \url{https://www.smooth-on.com/category/platinum-silicone/}.
\newblock Accessed: 2023-09-07.

\bibitem{c27}
S.~P. Boyd and L.~Vandenberghe, {\em Convex Optimization}.
\newblock Cambridge university press, 2004.

\bibitem{c19}
W.-T. Yang, H.~S. Stuart, and M.~Tomizuka, ``Mechanical modeling and optimal
  model-based design of a soft pneumatic actuator,'' in {\em IEEE International
  Conference on Soft Robotics (RoboSoft)}, pp.~1--7, IEEE, 2023.

\bibitem{yang2023pump}
W.-T. Yang, M.~Hirao, and M.~Tomizuka, ``Design, modeling, and parametric
  analysis of a syringe pump for soft pneumatic actuators,'' in {\em IEEE
  International Conference on Advanced Intelligent Mechatronics(AIM)},
  pp.~317--322, IEEE, 2023.

\bibitem{c29}
G.~Gerboni, A.~Diodato, G.~Ciuti, and A.~Cianchetti, Matteo~Menciassi,
  ``Feedback control of soft robot actuators via commercial flex bend
  sensors,'' {\em IEEE/ASME Transactions on Mechatronics}, vol.~22, no.~4,
  pp.~1881--1888, 2017.

\bibitem{yang2023control}
W.-T. Yang, B.~Kurkcu, M.~Hirao, L.~Sun, X.~Zhu, Z.~Zhang, G.~X. Gu, and
  M.~Tomizuka, ``Control of soft pneumatic actuators with approximated
  dynamical modeling,'' in {\em IEEE International Conference on Robotics and
  Biomimetics (ROBIO)}, pp.~1--8, IEEE, 2023.

\end{thebibliography}

\end{document}